\pdfoutput=1

\documentclass[11pt]{article}

\usepackage[]{EMNLP2022}

\usepackage{times}
\usepackage{latexsym}

\usepackage[T1]{fontenc}

\usepackage[utf8]{inputenc}

\usepackage{microtype}

\usepackage{inconsolata}

\usepackage{enumitem}
\usepackage{amstext}
\usepackage{booktabs}
\usepackage{graphicx}
\usepackage{mathtools}
\usepackage[ruled]{algorithm2e}
\usepackage{amsfonts,amssymb}
\usepackage{multirow}
\usepackage{caption}
\usepackage{subcaption}
\usepackage{array}

\title{ConReader: Exploring Implicit Relations in Contracts for\\ Contract Clause Extraction\thanks{\hspace{2mm} The work described in this paper is substantially supported by a grant from the Research Grant Council of the Hong Kong Special Administrative Region, China (Project Code: 14204418).}}

\author{
Weiwen Xu\raisebox{4pt}{\small $1$}, Yang Deng\raisebox{4pt}{\small $1$}, Wenqiang Lei\raisebox{4pt}{\small $2$}, Wenlong Zhao\raisebox{4pt}{\small $1$}, Tat-Seng Chua\raisebox{4pt}{\small $3$}, and Wai Lam\raisebox{4pt}{\small $1$} \\
\raisebox{4pt}{\small $1$}The Chinese University of Hong Kong \\
\raisebox{4pt}{\small $2$}Sichuan University \\
\raisebox{4pt}{\small $3$}National University of Singapore \\
{\tt \{wwxu,ydeng,wlam\}@se.cuhk.edu.hk}\\
{\tt \{wenqianglei,wenlzhao\}@gmail.com},
{\tt chuats@comp.nus.edu.sg}
}

\begin{document}
\maketitle
\begin{abstract}
    We study automatic \textbf{C}ontract \textbf{C}lause \textbf{E}xtraction (CCE) by modeling implicit relations in legal contracts.
    Existing CCE methods mostly treat contracts as plain text, creating a substantial barrier to understanding contracts of high complexity.
    In this work, we first comprehensively analyze the complexity issues of contracts and distill out three implicit relations commonly found in contracts, namely,
    1) \textit{Long-range Context Relation} that captures the correlations of distant clauses; 2) \textit{Term-Definition Relation} that captures the relation between important terms with their corresponding definitions; and 3) \textit{Similar Clause Relation} that captures the similarities between clauses of the same type.
    Then we propose a novel framework {\tt ConReader} to exploit the above three relations for better contract understanding and improving CCE.
    Experimental results show that {\tt ConReader} makes the prediction more interpretable and achieves new state-of-the-art on two CCE tasks in both conventional and zero-shot settings.\footnote{Our code is available at: \url{https://github.com/wwxu21/ConReader}}
\end{abstract}
\section{Introduction}
\begin{figure*}
    \centering
    \includegraphics[scale=0.46]{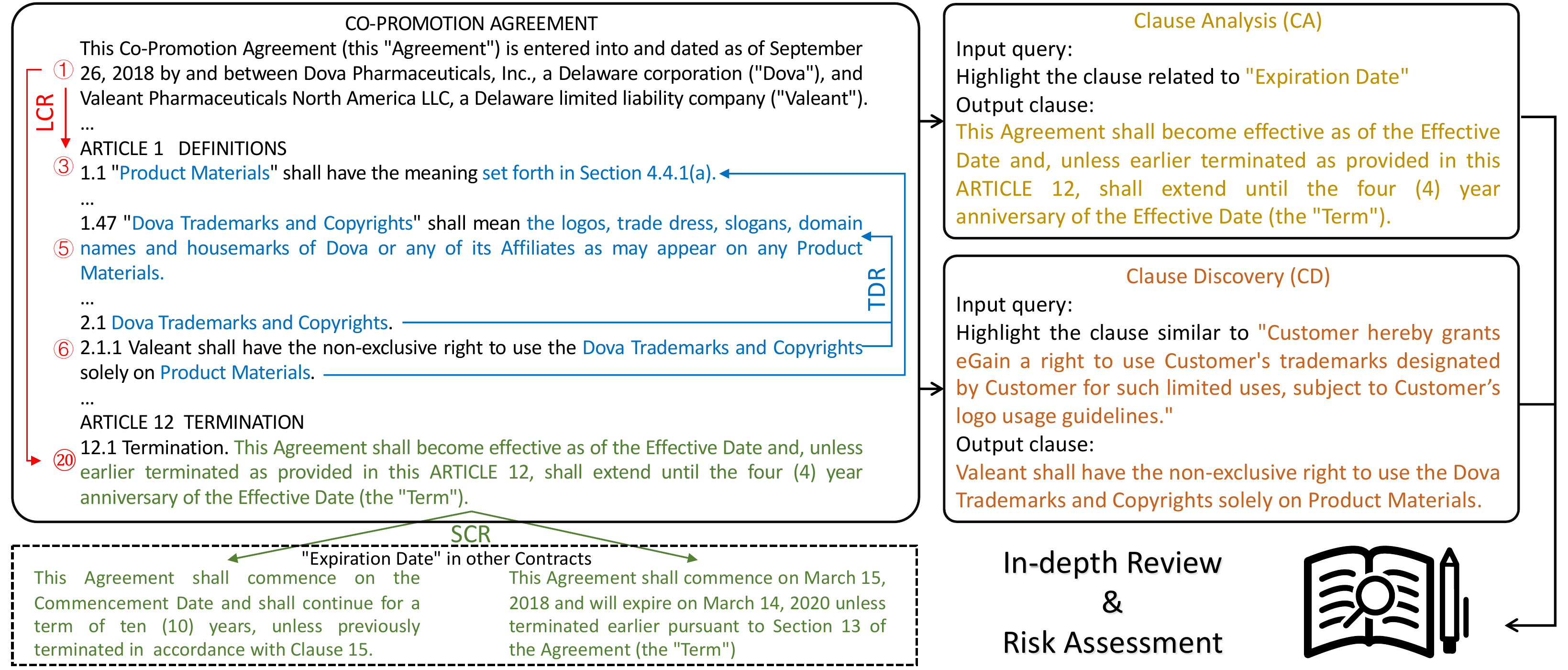}
    \caption{An overview of the contract structure and CCE process. The left half illustrates three implicit relations widely found in contracts. The right half shows two tasks of CCE.}
    \vspace{-5pt}
    \label{fig:example}
\end{figure*}
Legal Contract Review is a process of thoroughly examining a legal contract before it is signed to ensure that the content stated in the contract is clear, accurate, complete and free from risks.
A key component to this application is the Contract Clause Extraction (CCE), which aims to identify key clauses from the contract for further in-depth review and risk assessment.
Typically, CCE consists of two major tasks targeting different query granularities for real-life usages. They are \textit{Clause Analysis} (CA) and \textit{Clause Discovery} (CD) \footnote{CA refers to Contract Analysis in \citet{hendrycks2021cuad}. CD refers to Contract Discovery in \citet{borchmann-etal-2020-contract}.}, where CA aims to identify clauses that belong to a general clause type, while CD aims to identify clauses similar to a specific clause (depicted in Figure \ref{fig:example}).
CCE is both expensive and time-consuming as it requires legal professionals to manually identify a small number of key clauses from contracts with hundreds of pages in length \cite{hendrycks2021cuad}.
Therefore, there is a pressing need for automating CCE, which assists legal professionals to analyze long and tedious documents and provides non-professionals with immediate legal guidance.

The biggest challenge to automating CCE is the complexities of contracts.
In the literature, simply treating contracts as plain text, most pretrained language models perform poorly on CCE \cite{devlin2018bert,liu2019roberta}.
Some works try to simplify CCE from the perspective of contract structure. For example,
\citet{chalkidis2017extracting} assign a fixed extraction zone for each clause type and limit the clauses to be extracted only from their corresponding extraction zones.
\citet{hegel2021law} use visual cues of document layout and placement as additional features to understand contracts.
However, their local context assumption is not flexible and, more seriously, neglects more complicated relations inherent in the contracts.

In fact, as shown in Figure \ref{fig:example}, contracts are formal documents that typically follow a semi-structured organization. The body of a contract is usually organized into some predefined articles such as "Definitions" and "Terminations", where relevant clauses are orderly described inside. 
Different articles may hold different levels of importance. For example, the "Definitions" article is globally important because it clearly defines all important terms that would be frequently referenced, while other articles are sparsely correlated, holding local importance.
We attempt to decompose the complexities into a set of implicit relations, which can be exploited to better understand contracts.
Therefore,  as shown in Figure \ref{fig:example}, we identify three implicit relations to directly tackle the complexities from three aspects:

1) \textit{The implicit logical structure among distant text}: This is originated from the fact that a clause from one article may refer to clauses from distant articles. However, most pretrained language models (e.g. BERT) inevitably break the correlations among clauses because they have to split a contract into multiple segments for separate encoding due to the length limitation. Therefore, we define a \textbf{Long-range Context Relation} (LCR) to capture the relations between different segments to keep the correlations among clauses.

2) \textit{The unclear legal terms}: Legal terms need to be clearly and precisely declared to minimize ambiguity. Thanks to the "Definition" article, we can easily find the meaning of a particular term. Then the relation between each term and its definition is defined as \textbf{Term-Definition Relation} (TDR). The clarity of TDR allows consistent information flow by enhancing terms with semantics-rich definitions;

3) \textit{The ambiguity among clauses}:  It is usually hard to differentiate different types of clauses just from their text formats. For example, clauses of type "Expiration Date" and "Agreement Date" both show up as dates. 
It leads to the third relation defined as \textbf{Similar Clause Relation} (SCR). SCR captures the similarity of the same type of clauses across contracts. It enhances a clause's semantics with its unique type information and thus maintains the discrimination among different clause types.
Furthermore, LCR and TDR are two intra-contract relations while SCR is an inter-contract relation.

In light of the above investigations about the complexities of contracts, we propose a novel framework, {\tt ConReader}, to tackle two CCE tasks by exploiting the above three relations for better contract understanding.
Concretely, we reserve a small number of token slots in the input segments for later storage of the three kinds of relational information. To prepare intra-contract relations, including LCR and TDR, we get the segment and definition representations from pretrained language models. 
Regarding the inter-contract relation, i.e. SCR, since the size of SCR increases as the number of contracts increases, we are unable to enumerate all possible SCRs. Therefore, we enable input segments to interact with a \textit{Clause Memory} that stores recently visited clauses, where a clause retriever is adopted to retrieve similar clauses from the Clause Memory. 
Then, we enrich each segment by filling the reserved slots with context segments, relevant definitions, as well as retrieved similar clauses. Finally, a fusion layer is employed to simultaneously learn relevant information both from the local (i.e. within the segment) or global context (i.e. via implicit relations) for extracting the target clause.

To summarize, our main contributions are threefold:
\begin{itemize}[leftmargin=*,topsep=4pt]
\setlength{\itemsep}{0pt}
\setlength{\parskip}{0pt}
\setlength{\parsep}{0pt}
\item This work targets automatic CCE. We comprehensively analyze the complexity issues of modeling legal contracts and distill out three implicit relations, which have hardly been discussed before.
\item We propose a novel framework {\tt ConReader} to effectively exploit the three relations. It enables a more flexible relations modeling and reduces the difficulties in understanding contracts for better CCE.
\item Experimental results on two CCE tasks, namely CA and CD, show considerable improvements in both performance and interpretability.
\end{itemize}
\section{Framework}
\textbf{Overview}
We describe the problem definition for CCE via extractive Question Answering (QA) \cite{rajpurkar-etal-2016-squad}.
Let $\{\textbf{\textit{c}}_m\}_{m=1}^M$ be a contract in the form of multiple segments and $\textbf{\textit{q}}$ be a query either represented as a clause type in the CA task or a specific clause in the CD task. Our goal is to extract clauses $\{\textbf{\textit{y}}_k\}_{k=1}^K$ corresponding to the query. There may be multiple or no correct clauses and each clause is a text span in a particular segment denoted by its start and end index if existent.

\begin{figure}[t]
    \centering
    \includegraphics[scale=0.5]{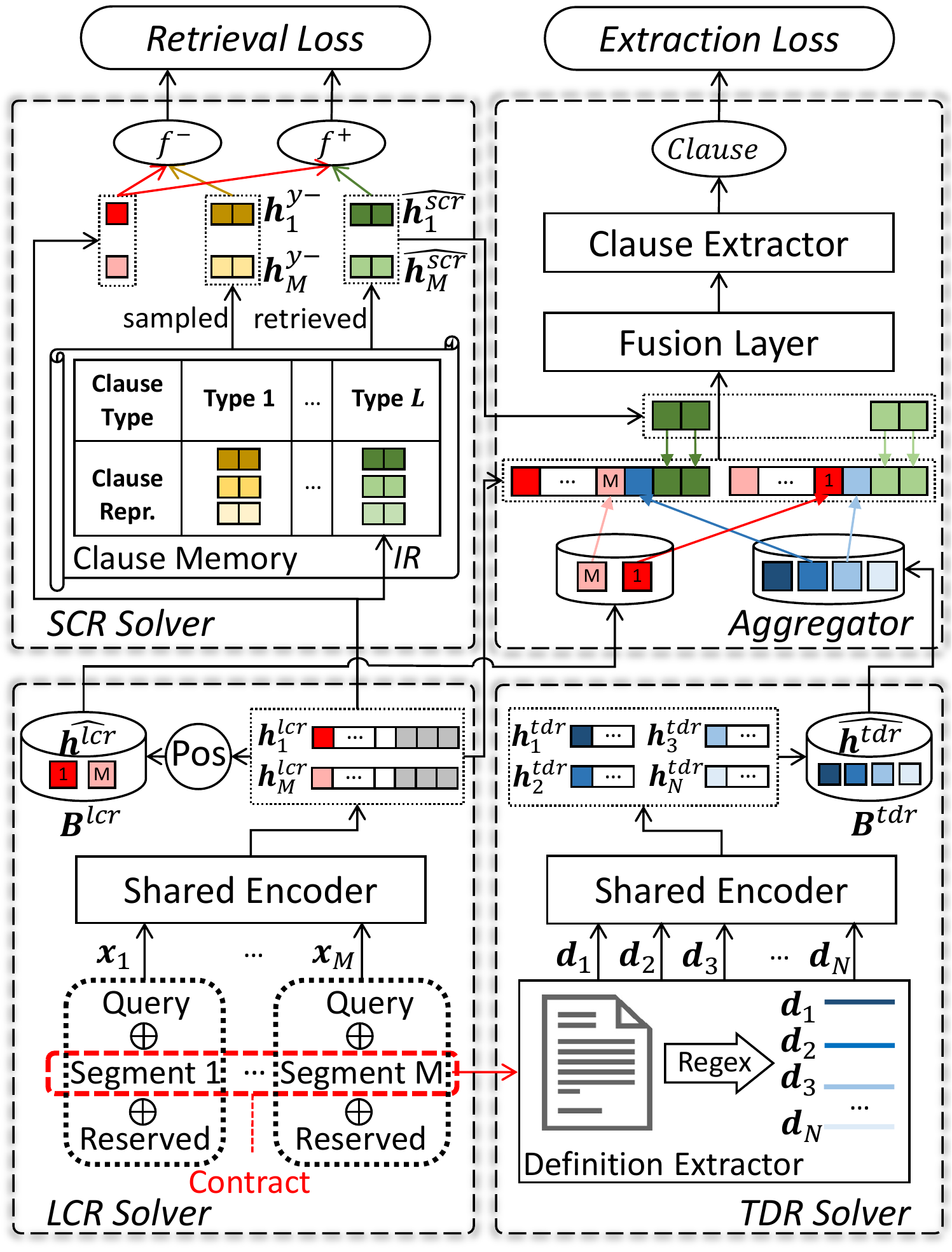}
    \caption{Overview of {\tt ConReader}. Three solvers are used to obtain relevant information and an Aggregator is used to fuse all information into text representations for semantic enrichment. IR denotes the retrieval process.}
    \label{fig:model}
\end{figure}

Figure \ref{fig:model} depicts the overview of {\tt ConReader}, which consists of four main components:
\begin{itemize}[leftmargin=*,topsep=4pt]
\setlength{\itemsep}{0pt}
\setlength{\parskip}{0pt}
\setlength{\parsep}{0pt}
\item \textit{LCR Solver} tackles LCR by encoding the wrapped segments $\{\textbf{\textit{x}}_m\}_{m=1}^M$ aware of the query $\textbf{\textit{q}}$ and the reserved slots $\textbf{\textit{r}}$ into hidden states $\{\textbf{\textit{h}}^{lcr}_m\}_{m=1}^M$, where the overall segment representations are stored in a segment bucket $\textbf{B}^{lcr}$.

\item \textit{TDR Solver} tackles TDR by encoding all definitions $\{\textbf{\textit{d}}_n\}_{n=1}^N$ from the contract into hidden states $\{\textbf{\textit{h}}^{tdr}_n\}_{n=1}^N$, where the overall definition representations are stored in a definition bucket $\textbf{B}^{tdr}$.

\item \textit{SCR Solver} tackles SCR by retrieving similar clause representations $\{\widehat{\textbf{\textit{h}}^{scr}_m}\}_{m=1}^M$ from a Clause Memory $\mathcal{M}$ according to a similarity function $f(\cdot,\cdot)$ between the segment and the stored clause.
\item \textit{Aggregator} enriches each segment representation with the three relational information for extracting the target clause. 
\end{itemize}

\subsection{Long-range Context Relation Solver}
\label{sec:lcr}
The goal of LCR Solver is to output all segment representations in a contract in the face of the length limitation of pretrained language models.
Meanwhile, to allow a flexible relation modeling in later Aggregator, we reserve some token slots for later storage of relational information before encoding.

Specifically, we concatenate each segment with the query and the reserved token slots to form the input sequence within the length limitation:
\begin{equation}
    \small
    \textbf{\textit{x}}_m=[{\tt [CLS]};\textbf{\textit{q}};{\tt [SEP]};\textbf{\textit{c}}_m;{\tt [SEP]};\textbf{\textit{r}}] \;\;\; m=1,...,M 
\end{equation}
where $[\cdot;\cdot]$ denotes the sequential concatenation, ${\tt [CLS]}$, ${\tt [SEP]}$ are special tokens at the beginning or in the middle of the two text. Note that the reserved token slots $\textbf{\textit{r}}$ are occupied with placeholders and only take a small portion of the entire sequence ($|\textbf{\textit{r}}| << 512$) such that they only slightly affect the efficiency. It does not matter which token is chosen as the placeholder since we would directly mask these slots such that they will not affect the hidden states of query and segment tokens as well as not receive gradient for update.

Then, we apply a RoBERTa encoder $\textbf{Enc}(\cdot)$ to get the hidden states for all input sequences: $\textbf{\textit{h}}^{lcr}_m=\textbf{Enc}(\textbf{\textit{x}}_m)$,
where $\textbf{\textit{h}}^{lcr}_m \in \mathbb{R}^{|\textbf{\textit{x}}_m|\times h}$, and $h$ is the hidden dimension. 
To reflect the order of different segments in a contract, we also add a segment positional embedding \cite{vaswani2017attention} to the hidden state $\textbf{\textit{h}}^{lcr}_{m,cls }$ at ${\tt [CLS]}$ to get the segment representation for each input segment:
\begin{equation}
    \small
    \widehat{\textbf{\textit{h}}^{lcr}_m} = \textbf{\textit{h}}^{lcr}_{m,cls} + \textbf{Pos}(m) 
\end{equation} 
where $\textbf{Pos}(\cdot)$ is a standard RoBERTa positional encoder. All segment representations are temporarily stored in a segment bucket $\textbf{B}^{lcr}=\{\widehat{\textbf{\textit{h}}^{lcr}_m}\}_{m=1}^M$.

\subsection{Term-Definition Relation Solver}
TDR Solver is responsible for providing the specific definitions for terms that may raise ambiguity. It can be observed in Figure \ref{fig:example} that definitions are well organized in the ``Definition" article. Therefore, we use regular expressions including some keywords like ``\textit{shall mean}", ``\textit{mean}" to automatically extract those definitions. Then, we prepare the definition inputs as :
\begin{equation}
\small
        \textbf{\textit{d}}_n=[{\tt [CLS]};\textbf{\textit{k}}_n;{\tt [SEP]};\textbf{\textit{v}}_n;{\tt [SEP]}] \;\;\; n=1,...,N 
\end{equation}
where each definition is presented in the form of key-value pair. Each key $\textbf{\textit{k}}_n$ denotes a legal term in the contract and the value $\textbf{\textit{v}}_n$ denotes its corresponding definition text. 
Then we apply the same RoBERTa encoder to encode these definitions into hidden states $\textbf{\textit{h}}^{tdr}_n$, where the hidden states $\textbf{\textit{h}}^{tdr}_{n,cls}$ at ${\tt [CLS]}$ are denoted as definition representations $\{\widehat{\textbf{\textit{h}}^{tdr}_n}\}_{n=1}^N$, which are temporarily stored in another definition bucket $\textbf{B}^{tdr}$.

\subsection{Similar Clause Relation Solver}
Since SCR is an inter-contract relation, we are unlikely to enumerate all possible clause pairs.
Therefore, we maintain a Clause Memory $\mathcal{M}$ to: (1) dynamically store clauses of all types; and (2) allow input segments to retrieve similar clauses according to a similarity function $f(\cdot,\cdot)$.
Details can be found in Algorithm \ref{algo}.

\begin{algorithm}[t]
    \small
    \caption{SCR Solver (training)}
    \LinesNumbered
    \label{algo}
	\KwIn{$\textbf{\textit{q}}$, $\{\textbf{\textit{c}}_m\}_{m=1}^M$, $\{\textbf{\textit{y}}_k\}_{k=1}^Y$;} 
    \KwOut{ $\{\widehat{\textbf{\textit{h}}^{scr}_m}\}_{m=1}^M$; }
    Initialize all parameters: $\mathcal{M}[l]=$ Queue(), $l=1,...,L$\;
    Get hidden states of segments $\{\textbf{\textit{h}}^{lcr}_m\}_{m=1}^M$ from Section \ref{sec:lcr} using $\textbf{\textit{q}}$ and $\{\textbf{\textit{c}}_m\}_{m=1}^M$\;
    Get clause type $l_q$ according to the query $\textbf{\textit{q}}$\;
	\texttt{// retrieve clauses}\;
	\For{segment $m = 1, 2, \ldots , M$}{
		Retrieve a similar clause $\widehat{\textbf{\textit{h}}^{scr}_m}$  for each segment via Equation (\ref{equ:retrieval})\;
	}
	\texttt{// Update clause memory}\;
	\For{extractable clause $k = 1, 2, \ldots , K$}{
        Get clause representation $\textbf{\textit{h}}^{y_k}$ via Equation (\ref{equ:clause})\;
        \If{memory partition $\mathcal{M}[l_q]$ is full}{
        Remove the earliest clause representation\;
        }

        En-queue $\textbf{\textit{h}}^{y_k}$ to $\mathcal{M}[l_q]$\;
    }
\end{algorithm}

\paragraph{Dynamic Update of $\mathcal{M}$}
During training, we assume each query $\textbf{\textit{q}}$ implies a particular clause type $l_q$ (the query of CA itself is a clause type, while the query of CD belongs to a clause type), where we have $L$ clause types in total. 
Initially, $\mathcal{M}$ allocates the same memory space of size $|\mathcal{M}|$ for each clause type to store the corresponding clause representations. 
Suppose that we get $\textbf{\textit{h}}^{lcr}_{m}$ from LCR Solver for $\textbf{\textit{x}}_m$ and there is a clause $\textbf{\textit{y}}$ of type $l_q$ corresponding to the given query $\textbf{\textit{q}}$ inside $\textbf{\textit{x}}_m$. We denote its clause representation $\textbf{\textit{h}}^{y}$ as the concatenation of its start and end token representations:
\begin{equation}
\label{equ:clause}
    \textbf{\textit{h}}^{y}=[\textbf{\textit{h}}^{lcr}_{m,s}:\textbf{\textit{h}}^{lcr}_{m,e}] \in \mathbb{R}^{2h}
\end{equation}
where $[\cdot:\cdot]$ denotes vector concatenation, and $s$ and $e$ are the start and end index of $\textbf{\textit{y}}$ inside $\textbf{\textit{x}}_m$. When encountering such clause, we add $\textbf{\textit{h}}^{y}$ to its corresponding memory partition $\mathcal{M}[l_q]$. If the memory partition is full, we follow the first-in first-out (FIFO) principle to remove the earliest clause representation stored in $\mathcal{M}[l_q]$ to make room for the new one, such that the clause representations stored are always up-to-date.

\paragraph{Retrieve Clauses from $\mathcal{M}$}
When asking to identify clause of type $l_q$, we allow each input segment to retrieve a similar clause from the Clause Memory. The retrieved clause would imply the semantic and contextual information of this type of clauses in other contracts, facilitating the extraction of the same type of clauses in the current contract.

Specifically, given the hidden states of the input sequence  $\textbf{\textit{h}}^{lcr}_{m}$ with a query $\textbf{\textit{q}}$ of type $l_q$ as well as the Clause Memory $\mathcal{M}$, we limit the retrieval process only in the corresponding memory partition $\mathcal{M}[l_q]$ during training to retrieve truly similar (i.e. of the same type) clauses that provide precise guidance on clause extraction in the current contract. The retriever is implemented as a similarity function $f(\cdot,\cdot)$:
\begin{equation}
\label{equ:retrieval}
    \widehat{\textbf{\textit{h}}^{scr}_m}=\mathop{\arg\max}\nolimits_{\textbf{\textit{h}}^{y} \in \mathcal{M}[l_q]} f(\textbf{\textit{h}}^{lcr}_{m,cls},\textbf{\textit{h}}^{y})
\end{equation}
where $f(\textbf{\textit{h}}^{lcr}_{m,cls},\textbf{\textit{h}}^{y})=\cos{(\textbf{\textit{h}}^{lcr}_{m,cls}\textbf{\textit{W}}^{lcr},\textbf{\textit{h}}^{y}\textbf{\textit{W}}^y)}$, $\textbf{\textit{W}}^{lcr} \in \mathbb{R}^{h \times h}$ and $\textbf{\textit{W}}^{y} \in \mathbb{R}^{2h \times h}$ are parameters to project $\textbf{\textit{h}}^{lcr}_{m,cls},\textbf{\textit{h}}^{y}$ to the same space.

To make the retriever trainable such that it can learn to capture the common characteristics of the same type of clauses, we introduce a Retrieval Loss $\mathcal{L}_{r}$ to minimize a contrastive learning loss function \cite{hadsell2006dimensionality}, where a negative clause $\textbf{\textit{h}}^{y-} \in \mathcal{M}\setminus \mathcal{M}[l_q]$ is randomly sampled:
\begin{equation}
\small
    \mathcal{L}_{r}=\sum_{m=1}^M \max(0, 1-f(\textbf{\textit{h}}^{lcr}_{m,cls},\widehat{\textbf{\textit{h}}^{scr}_m}) + f(\textbf{\textit{h}}^{lcr}_{m,cls},\textbf{\textit{h}}^{y-})) 
\end{equation}

\subsection{Aggregator}
\label{sec:aggr}
After obtaining relational information from corresponding relation solvers, we fill all these representations into the reserved token slots and allow the new segment sequence to automatically learn three implicit relations via a fusion layer.

For LCR and TDR, not all segment or definition representations in the corresponding buckets are necessary for each input segment as they may be repeated (i.e. LCR) or out of segment scope (i.e. TDR). Therefore, for the $m$-th input segment, we remove the repeated segment representation (i.e. $\widehat{\textbf{\textit{h}}^{scr}_m}$) and only consider the definition representations whose terms appear in this segment:
\begin{equation}
\begin{split}
\small
    \textbf{B}^{lcr}_m&=\textbf{B}^{lcr} \setminus \widehat{\textbf{\textit{h}}^{scr}_m}\\ \textbf{B}^{tdr}_m&=\{\widehat{\textbf{\textit{h}}^{tdr}_n}\;|\;\textbf{\textit{d}}_n \;\text{in}\;\textbf{\textit{c}}_m, n\in[1,N]\}
\end{split}
\end{equation}

For SCR, each segment is paired with one clause representation retrieved. Then after filling all corresponding representations into the reserved slots, we get the final hidden state $\textbf{\textit{h}}_m$ for each segment:
\begin{equation}
    \textbf{\textit{h}}_m=[\textbf{\textit{h}}^{lcr}_{m,cls:sep2};\textbf{B}^{lcr}_m;\widehat{\textbf{\textit{h}}^{scr}_m};\textbf{B}^{tdr}_m] 
\end{equation}
where $\textbf{\textit{h}}^{lcr}_{m,cls:sep2}$ are the hidden states ranging from $\tt[CLS]$ to the second $\tt[SEP]$ in $\textbf{\textit{h}}^{lcr}_{m}$. Note that we do not set a specific size of reserved slots for each relation, but only assure that the total size should not exceed $|\textbf{\textit{r}}|$. The reserved slots taken by these representations are unmasked to enable calculation and gradient flow.
Then $\textbf{\textit{h}}_m$ would pass a fusion layer to  automatically learn the three implicit relations:
\begin{equation}
    \textbf{\textit{o}}_m=\textbf{Fusion}(\textbf{\textit{h}}_m)
\end{equation}
where $\textbf{Fusion}(\cdot)$ is a standard RoBERTa layer with randomly initialized parameters and $\textbf{\textit{o}}_m$ is the relation-aware hidden states for the $m$-th segment. We use $\textbf{\textit{o}}_m$ to extract clause:
\begin{equation}
\begin{split}
\small
    P_s(m)&=\text{softmax}(\textbf{\textit{o}}_m \textbf{\textit{W}}^s)\\
    P_e(m)&=\text{softmax}(\textbf{\textit{o}}_m \textbf{\textit{W}}^e) 
\end{split}
\end{equation}
where $P_s(m)$ and $P_e(m)$ denote the probabilities of a token being the start and end positions respectively. $\textbf{\textit{W}}^s, \textbf{\textit{W}}^e \in \mathbb{R}^{h \times 1}$ are corresponding parameters. The Extraction Loss $\mathcal{L}_{e}$ is defined as the cross-entropy between the predict probabilities and the ground-truth start and end positions respectively. 

\subsection{Training \& Prediction}
\paragraph{Training}
During training, we assume that the clause type for each input query is available and follow {\tt ConReader} to get $\mathcal{L}_{r}$ and $\mathcal{L}_{e}$, where the final training objective is the summation of them $\mathcal{L}=\mathcal{L}_{r} + \mathcal{L}_{e}$.
If no clauses can be extracted given the current query, we set both the start and end positions to 0 (i.e. $\tt [CLS]$).

\paragraph{Prediction}
At the prediction time, we may encounter zero-shot scenarios where the clause types are out-of-scope of the existing $L$ types and, more seriously, CD essentially does not provide the clause type for each query clause.
This would stop {\tt ConReader} from generalizing to these scenarios as we are unable to indicate which memory partition of $\mathcal{M}$ for retrieval. To address this limitation, we allow the retrieval to be performed in the entire clause memory ( the condition in Equation \ref{equ:retrieval} would be replaced to $\textbf{\textit{h}}^{y} \in \mathcal{M}$) since the retriever has already learned to effectively capture the common characteristics of similar clauses. To deal with the extraction of multiple clauses, we follow \citet{hendrycks2021cuad} to output top $T$ clauses according to $P_s(m)_i \times P_e(m)_j$ in the contract, where $0\leq i\leq j\leq|\textbf{\textit{x}}_m|$ denote positions in $\textbf{\textit{x}}_m$.

\section{Experimental Settings}
We conduct experiments on two CCE tasks, namely CA and CD, in two settings: 
(1) the conventional setting where the clauses in the training and test sets share the same clause types; and (2) a more difficult zero-shot setting where the clause types differ substantially for training and test set.

\paragraph{Datasets}
To implement {\tt ConReader} on CA and CD in both settings, we combine two datasets which originally only tackle one of the tasks:
\begin{itemize}[leftmargin=*,topsep=4pt]
\setlength{\itemsep}{0pt}
\setlength{\parskip}{0pt}
\setlength{\parsep}{0pt}
\item  \textbf{CUAD} \cite{hendrycks2021cuad} is proposed to only tackle CA. It carefully annotates 41 types of clauses that warrant review. CUAD provides CA datasets for both training and test.
\item  \textbf{Contract Discovery} \cite{borchmann-etal-2020-contract} is proposed to only tackle CD. It annotates 21 types of clauses substantially different from CUAD and applies a repeated sub-sampling procedure to pair two clauses of the same type as a CD example. However, since the legal annotation is expensive, it only provides development and test sets.
\end{itemize}

For CA, we  use the training set of CUAD to train a {\tt ConReader} model. We evaluate it on the test set of CUAD for the conventional setting and on the development and test sets of Contract Discovery for the zero-shot setting. 
For CD, since we now have a training set from CUAD, we apply the same supervised extractive QA setting, where one clause is supposed to be extracted conditioned on the query clause instead of original unsupervised sentence matching formulation.
Similar to \citet{borchmann-etal-2020-contract}, we sub-sample k (k = 5 in our work) clauses for each clause type and split them into k - 1 seed clauses and 1 target clause. 
Then, we pair each of the seed clauses with the contract containing the target clause to form k - 1 CD examples.
By repeating the above process, we can finally get the CD datasets for both training and evaluation.
Similar to CA, we train another model for CD and evaluate it in two settings.
Details of data statistics can be found in Appendix \ref{sec:data}.

\paragraph{Evaluation Metrics}
Following \citet{hendrycks2021cuad}, we use Area Under the Precision-Recall curve (AUPR) and Precision at 80\% Recall (P@0.8R) as the major evaluation metrics for CA.
In CUAD, an extracted clause is regarded as true positive if the Jaccard similarity coefficient between the clause and the ground truth meets a threshold of 0.5 \cite{hendrycks2021cuad}. While in Contract Discovery, it tends to annotate longer clauses with some partially related sentences (examples can be found in Appendix \ref{sec:anno}). Therefore, we also regard an extracted clause as true positive if it is a sub-string of the ground truth.
For CD, we use AUPR and Soft-F1 to conduct a more fine-grained evaluation in terms of words \cite{borchmann-etal-2020-contract}.

\paragraph{Baseline Methods}
We compare with several recently published methods, including: 1) Rule-based or unsupervised contract processing models: Extraction Zone \cite{chalkidis2017extracting} and Sentence Match \cite{borchmann-etal-2020-contract}; 2) Strong pretrained language models: BERT \cite{devlin2018bert}, RoBERTa \cite{liu2019roberta}, ALBERT \cite{lan2020albert}, DeBERTa \cite{he2020deberta} and RoBERTa+PT that pretrained on 8GB contracts \cite{hendrycks2021cuad}; and 3) Models tackling long text issue: Longformer \cite{beltagy2020longformer}, and Hi-Transformer \cite{wu-etal-2021-hi}.

\paragraph{Implementation Details}
We apply our framework on top of two model sizes, namely, RoBERTa-base (12-layer, 768-hidden, 12-heads, 125M parameters) and RoBERTa-large (24-layer, 1024-hidden, 16-heads, 355M parameters) from Huggingface\footnote{\url{https://github.com/huggingface/transformers}}.
The reserved slots size $|\textbf{\textit{r}}|$ is set to 30 such that most of the relational information can be filled in. The size of Clause Memory $|\mathcal{M}|$ for each partition is 10. In prediction, we follow \citet{hendrycks2021cuad} to output top $T=20$ clauses. 
Recall that the query of CD is a clause, which is much longer than a clause type. We set the max query length for CA and CD to be 64 and 256 respectively. The max sequence length is 512 for both models in two tasks.
We follow the default learning rate schedule and dropout settings used in RoBERTa. We use AdamW~\cite{loshchilov2018decoupled} as our optimizer.
We use grid search to find optimal hyper-parameters, where the learning rate is chosen from \{1e-5,5e-5,1e-4\}, the batch size is chosen from \{6,8,12,16\}.

We additionally introduce 1.7M and 7M parameters to implement the clause retriever $f(\cdot,\cdot)$ and fusion layer \textbf{Fusion} in {\tt ConReader}. Comparing to RoBERTa, their sizes are almost negligible, and hardly affect the speed. All experiments are conducted on one Titan RTX card.

\section{Results}
\begin{table}
    \centering
    \small
    \setlength{\tabcolsep}{0.6mm}{
    \begin{tabular}{l|c||cccc}
    \toprule
    \multirow{2}{*}{Methods} & \multirow{2}{*}{\#Params} & \multicolumn{2}{c}{CA} &\multicolumn{2}{c}{CD}\\
    \cmidrule(lr){3-4}\cmidrule(lr){5-6}
    & & AUPR  & P@0.8R & AUPR  & Soft-F1  \\ \midrule
    Extraction Zone         & -     & 13.2 & 0    & -    & -   \\
    Sentence Match          & -     & -    & -    & 10.2  & 34.2\\
    BERT-b               & 109M  & 31.2 & 10.6 & 20.7 & 55.1\\
    ALBERT-b             & 11M   & 36.0 & 13.1 & 23.4 & 59.1\\
    RoBERTa-b            & 125M  & 43.2 & 32.2 & 29.6 & 63.8\\
    RoBERTa+PT-b         & 125M  & 45.2 & 34.1 & -    & -   \\
    Longformer-b         & 149M  & 45.8 & 0    & 22.4 & 54.6\\
    Hi-Trans.-b     & 295M  & 44.0 & 33.3 & 31.2 & 64.2\\ \midrule
    {\tt ConReader}-b    & 134M  & \textbf{47.2} & \textbf{38.7} & \textbf{33.5} & \textbf{66.1}\\\midrule \midrule
    BERT-l              & 335M  & 33.4 & 12.4 & 22.5 & 58.8 \\
    ALBERT-xxl          & 223M  & 38.4 & 31.0 & -    & -     \\
    RoBERTa-l           & 355M  & 47.4 & 38.9 & 34.6 & 67.5      \\
    DeBERTa-xl          & 750M  & 47.8 & 44.0 & -    & -     \\\midrule
      {\tt ConReader}-l & 364M  & \textbf{49.1} & \textbf{44.2} & \textbf{35.0} & \textbf{68.1}\\
\bottomrule
    \end{tabular}}
    \caption{Model Comparisons in the conventional setting. Results are divided into two groups according to their parameters size (-b denotes -base, -l denotes -large).}
    \label{tab:conventional}
\end{table}

\paragraph{Conventional Setting}  Table \ref{tab:conventional} shows the results of CA and CD in the conventional setting. Among base-size models, {\tt ConReader}-base significantly improves over all previous methods on both tasks, where it surpasses the RoBERTa-base by 4.0 and 3.9 AUPR respectively. 
Among large-size models, {\tt ConReader}-large can exceed RoBERTa-large by 1.7 AUPR and 5.3 P@0.8R on CA and achieves the new state-of-the-art. 
Such a large improvement on P@0.8R would make the model less likely to miss important clauses that may cause huge losses, which is especially beneficial in the legal domain.
Notably, {\tt ConReader}-large also exceeds DeBERTa-xlarge by 1.3 AUPR with less than half of its parameters (364M vs 750M), demonstrating the effectiveness of our framework.

Additionally, there are several notable observations:
1) As the queries in CD are clauses, they are more diverse than the 41 queries of CA, making it a more difficult CCE task.
2) We find that {\tt ConReader}-base outperforms RoBERTa+PT-base. This implies that explicitly modeling the complexities of the contracts is more valuable than learning from the in-domain data in an unsupervised manner.
3) The improvements of the models designed for long text (Longformer and Hi-Transformer) are less significant than {\tt ConReader}. 
It suggests that there are more sophisticated issues in contracts other than long text. In addition, Longformer favors Precision than Recall, causing P@0.8R to be 0 in CA and low performance in CD. Such a characteristic is not suitable for CCE as it has lower tolerance to miss important clauses.

\begin{table}
\setlength{\abovecaptionskip}{5pt}   
\setlength{\belowcaptionskip}{5pt}
    \centering
    \small
    \begin{tabular}{lcccc}
    \toprule
    \multirow{2}{*}{Methods} & \multicolumn{2}{c}{CA} &\multicolumn{2}{c}{CD}\\
    \cmidrule(lr){2-3}\cmidrule(lr){4-5}
     & Dev  &  Test & Dev  &  Test  \\ \midrule
      BERT-base        &3.7 &4.7     & 6.1 &7.5 \\
      RoBERTa-base           & 13.7  & 14.8 & 10.7 & 11.2\\
      Longformer-base        & 3.2 & 3.8    &2.6  &2.9 \\
      Hi-Transformer-base    & 12.9 & 13.8 & 10.5 & 10.7\\ \midrule
      {\tt ConReader}-base   & \textbf{14.8} & \textbf{15.9} & \textbf{11.9} & \textbf{12.4}\\\bottomrule
    \end{tabular}
    \caption{AUPR in the zero-shot setting.}
    \label{tab:zero}
\end{table}

\paragraph{Zero-shot Setting}
In Table \ref{tab:zero}, we show the results of CCE in the zero-shot setting, where users may look beyond the 41 types of clauses annotated in \citet{hendrycks2021cuad} for their particular purposes. We can observe that:
1) All models suffer from a great performance drop in both tasks due to the label discrepancy between training and evaluation, which highlights the challenge of CCE in the zero-shot setting.
2) Though Longformer-base performs well in the conventional setting, it is less competitive against RoBERTa-base in the zero-shot setting. 
We conjecture that it sacrifices the attention complexity for encoding longer text, which is hard to capture the semantic correlations never seen before in the zero-shot setting.
3) {\tt ConReader}-base achieves superior generalization ability in the zero-shot setting. This is because the three implicit relations widely exist in contracts, which are not restricted to a particular clause type.

\begin{table}
\setlength{\abovecaptionskip}{5pt}   
\setlength{\belowcaptionskip}{5pt}
    \centering
    \small
    \setlength{\tabcolsep}{1mm}{
    \begin{tabular}{lcccc}
    \toprule
    \multirow{2}{*}{Methods} & \multicolumn{2}{c}{CA} &\multicolumn{2}{c}{CD}\\
    \cmidrule(lr){2-3}\cmidrule(lr){4-5}
     & AUPR  & P@0.8R & AUPR  & Soft-F1  \\ \midrule
      {\tt ConReader}-base   & \textbf{47.2} & \textbf{38.7} & \textbf{33.5} & \textbf{66.1}\\
      -~w/o LCR              & 46.4 & 36.3 & 33.0 & 65.7\\
      -~w/o TDR             & 44.1 & 34.8 & 32.8 & 65.9\\
      -~w/o SCR             & 45.3 & 35.7 & 32.0 & 65.9\\ \bottomrule
    \end{tabular}}
    \caption{Ablation studies in the conventional setting.}
    \label{tab:ablation}
\end{table}

\paragraph{Ablation Study}
To investigate how each relation type contributes to CCE, we conduct an ablation study by ablating one component of {\tt ConReader} in each time, which is shown in Table \ref{tab:ablation}. For clarity, discarding LCR Solver means that we do not fuse segment representations in Aggregator but we still split a contract into segments for separate encoding. 
1) Discarding LCR Solver would slightly degrade the performance. 
Since LCR only appeals to a small number of clauses that require distant interactions, it has little benefit to the clauses that require interaction within a segment. This limits LCR in contributing to CCE.
2) The ablation study in terms of TDR shows that definition information actually improves CCE. It enhances the representations of terms with specific explanations, which makes them less ambiguous and thus allows consistent information flow. 
3) Discarding SCR Solver and the Retrieval Loss would also cast a serious impact on the results, especially on CD. Since the Retrieval Loss is a learning objective concerning the semantics of clauses, it benefits CD by alleviating the difficulty in understanding the query semantics.
As a result, LCR, SCR, and TDR should all be taken into consideration for building reliable CCE models.

\section{Further Analyses}
\begin{table}
\setlength{\abovecaptionskip}{2pt}   
\setlength{\belowcaptionskip}{2pt}
	\begin{subtable}[t]{3in}
        \centering
        \small
        \setlength{\tabcolsep}{1mm}{
        \begin{tabular}{l||cccc}
            \toprule
            Dataset     & \#~Contract   & \#~Definition   & F1@D      & Acc@C\\ \midrule
            Train       & 290       &  4256     & 97.2      & 75.2      \\
            Test        & 65        & 670       & 97.7      & 81.4      \\ 
            Total       & 355       & 4926      & 97.3      & 76.5      \\\bottomrule
        \end{tabular}}
        \caption{Definition statistics. F1@D denotes F1 on the definition level and Acc@C denotes the accuracy on the contract level.}
	\end{subtable}
	\begin{subtable}[t]{3in}
	    \centering
	    \small
        \begin{tabular}{l||ccc}
            \toprule
            Tasks & RoBERTa-base   & + Auto   & + Manual \\ \midrule
            CA    & 43.2      & 45.6      & 46.0 \\
            CD    & 29.6      & 31.5      & 31.8 \\\bottomrule
        \end{tabular}
         \caption{Model performance (AUPR) when enhancing RoBERTa-base either with automatically extracted (+Auto) or manually annotated (+Manual) definitions.}
        \end{subtable}
    \caption{Analysis of TDR Solver.}
    \label{tab:tdr}
\end{table}
\paragraph{Analysis of TDR Solver}
The quality of extracted definitions is of vital importance as it directly determines the effectiveness of definition representations. Therefore, to check the quality of our automatically extracted definitions, we compare them with ground-truth definitions annotated by us in CUAD.
The statistics of ground-truth definitions and the quality of automatically extracted definitions are shown in Table \ref{tab:tdr}.
Specifically, more than half of the contracts contain definitions (290 / 408 for training, 65 / 102 for test), where our rule-based extraction can correctly extract definitions for most of them. In addition, the results in Table \ref{tab:tdr} (b) show our extracted definitions (+Auto) are capable of improving the ability of baseline models to extract clauses by enhancing the representations of legal terms and their benefits are almost the same as the ground-truth definitions (+Manual).

\paragraph{Analysis of SCR Solver}
To examine in depth the effect of SCR Solver, we implement several variants from the perspectives of gathering similar clauses (Access) and maintaining the Clause Memory (Update).
As shown in Table \ref{tab:scr}, for Access, we evaluate two variants by randomly selecting a clause representation from the corresponding memory partition (w/ Random $\mathcal{M}[l_q]$) or retrieving the most similar one from the entire memory (w/ Retrieved $\mathcal{M}$ ). Since the first variant selects a truly positive example (of the same type) to train the Retrieval Loss, the performance only drops marginally comparing to our default design. While the second variant is less effective since it cannot guarantee the retrieval of a positive example, which imposes a distracting signal in the Retrieval Loss. For Update, we replace our FIFO update strategy with random update (w/ Random Update) or stopping update when memory is full (w/o Update). The first variant can also partially keep the clause representations update, while the second variant cannot, causing it to be less effective due to poor clause representations. Overall, our default design for SCR Solver is more effective than those variants.

\begin{table}[t]
    \centering
    \small
    
    \begin{tabular}{p{40mm}cc}
    \toprule
    Methods                            & CA    & CD\\\midrule
    RoBERTa-base & 43.2  & 29.6 \\
    w/ SCR Solver (Default) & 45.1  & 31.8 \\
    \midrule
    \multicolumn{3}{c}{ \em \small Access} \\\midrule
    w/ Random $\mathcal{M}[l_q]$        & 44.9      & 31.6  \\
    w/ Retrieved $\mathcal{M}$          & 44.5      & 31.0\\\midrule
    \multicolumn{3}{c}{ \em \small Update} \\\midrule
    w/ Random Update                    & 45.0  & 31.5 \\
    w/o Update                          & 44.5  & 30.6\\\bottomrule
    \end{tabular}
    \caption{AUPR on different variants of SCR Solver.}
    \label{tab:scr}
\end{table}

\begin{figure}[t]
\setlength{\abovecaptionskip}{2pt}   
\setlength{\belowcaptionskip}{2pt}
    \centering
    \includegraphics[scale=0.3]{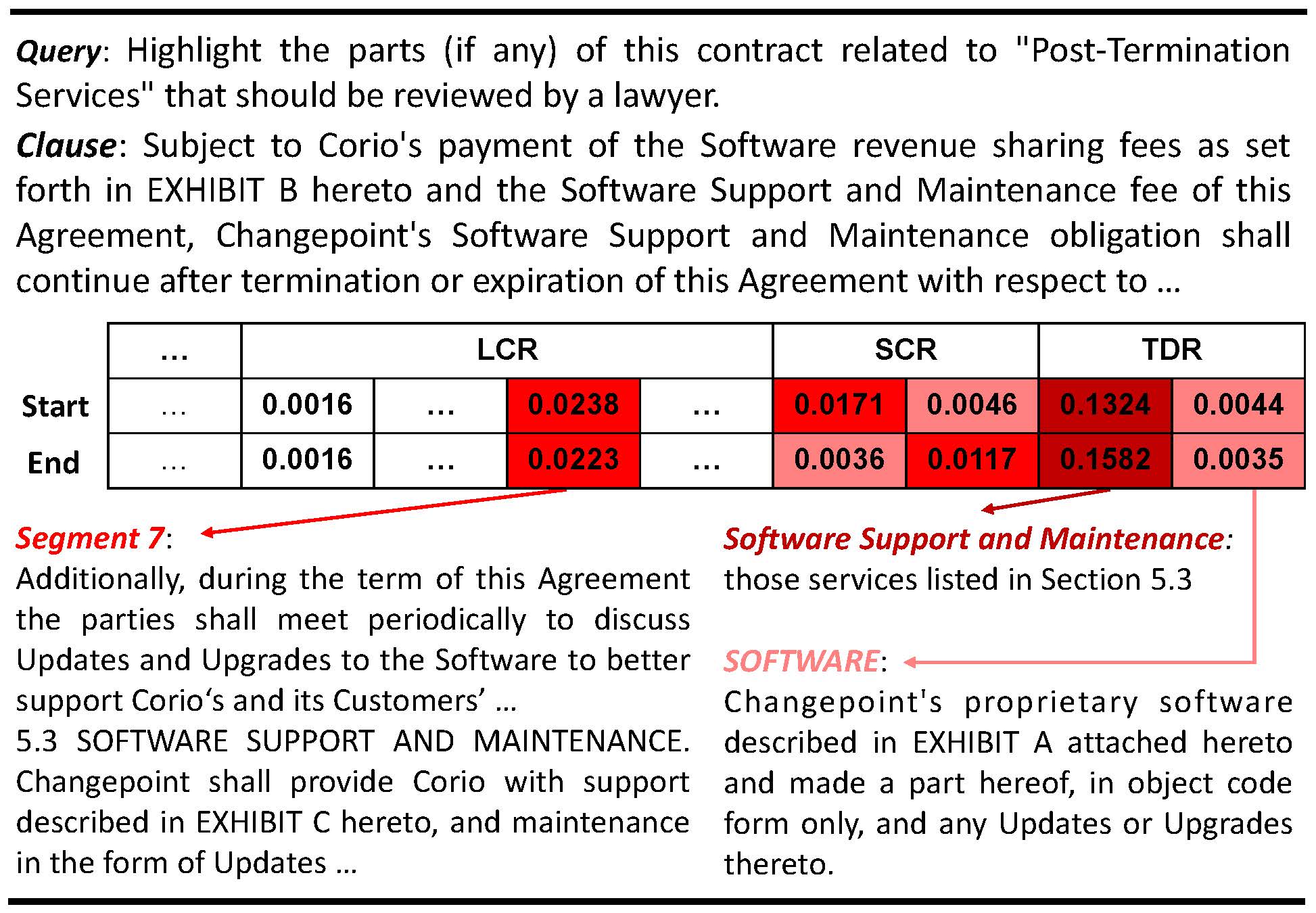}
    \caption{Case study of the attention distribution of a clause over its relevant information. }
    \label{fig:case}
\end{figure}
\paragraph{Case Study}
Figure \ref{fig:case} shows the attention distribution of the start and end tokens of the ground-truth clause over the reserved slots. It provides the interpretability that {\tt ConReader} can precisely capture the relevant relations with high attention probability. 
For example, it indicates that there is an important cue ("\textit{Section 5.3}") in the No.7 segment. It provides the detailed explanation of relevant terms ("\textit{Software Support and Maintenance}" and ``\textit{SOFTWARE}") that mentioned in this clause. In addition, the start and end tokens also exhibit high correlations with corresponding SCR start and end representations, showing that similar clauses can help determine the exact clause location.

\begin{figure}[t]
    \centering
    \includegraphics[scale=0.25]{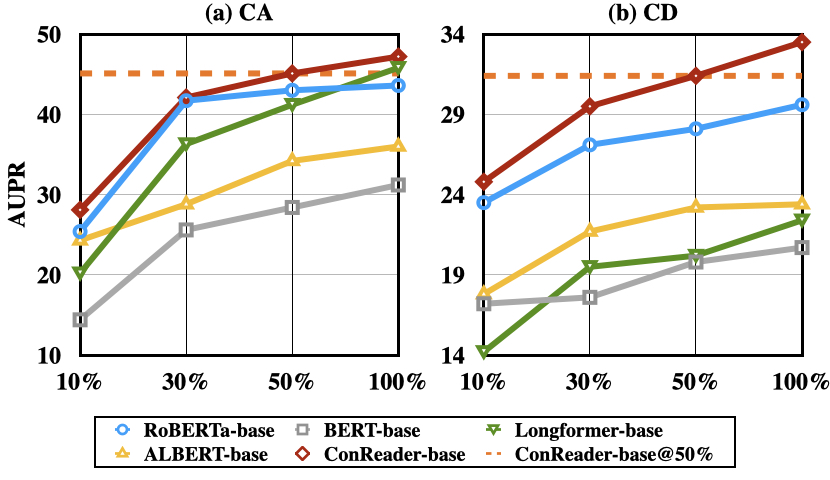}
    \caption{Performance (AUPR) w.r.t. training data size.}
    \label{fig:low}
\end{figure}

\paragraph{Effect of Training Data Size}
We simulate low-resource scenarios by randomly selecting 10\%, 30\%, 50\%, and 100\% of the training data for training CCE models and show the comparison results among various methods.
The performance trends  are visualized in Figure ~\ref{fig:low}.
In general, {\tt ConReader}-base makes an consistent improvement on different data sizes.
Impressively, it can yield an absolute increase of 14 AUPR on CA by increasing the training volume from 10\% to 30\%.
{\tt ConReader}-base with 50\% of the training data ({\tt ConReader}-base@50\%) can reach or almost exceed the performance of other approaches trained on 100\% training data on both CA and CD. 
These results shall demonstrate the great value of {\tt ConReader} in maintaining comparable performance and saving annotation costs at the same time.
Meanwhile, the performance trends of the two tasks indicate that there is still a lot of room for improvement, suggesting that the current bottleneck is the lack of training data.
According to the above analysis, we do believe that applying {\tt ConReader} can still achieve stronger results than textual-input baselines (e.g. RoBERTa) when more data is available and therefore, reduce more workload of the end users.

\section{Related Work}
\paragraph{Contract Review}
Earlier works start from classifying lines of contracts into predefined labels, where handcrafted rules and simple machine learning methods are adopted \cite{curtotti-mccreath-2010-corpus}. Then, some works take further steps to analyze contracts in a fine granularity, where a small set of contract elements are supposed to be extracted, including named entities \cite{chalkidis2017extracting}, parties' rights and obligations \cite{funaki-etal-2020-contract}, and red-flag sentences \cite{leivaditi2020benchmark}. They release corpora for automatic contract review, allowing neural models to get surprising performance \cite{chalkidis2017deep,chalkidis2019neural}. Recently, studies grow increasing attention on CCE to extract clauses, which are complete units in contracts, and carefully select a large number of clause types worth human attention \cite{borchmann-etal-2020-contract,sigir21-cross-cee,hendrycks2021cuad}. 
Due to the repetition of contract language that new contracts usually follow the template of old contracts \cite{simonson-etal-2019-extent}, existing methods tend to incorporate structure information to tackle CCE.
For example, \citet{chalkidis2017extracting} assign a fixed extraction zone for each clause type and limit the clauses to be extracted from corresponding extraction zones.
\citet{hegel2021law} leverage visual cues such as document layout and placement as additional features to better understand contracts.

\paragraph{Retrieval \& Memory}
Retrieval from a global memory has shown promising improvements to a variety of NLP tasks as it can provide extra or similar knowledge. One intuitive application is the open-domain QA, where it intrinsically necessitates retrieving relevant knowledge from outer sources since there is no supporting information at hand \cite{chen-etal-2017-reading,karpukhin-etal-2020-dense,xu-etal-2021-exploiting-reasoning,xu-etal-2021-dynamic}. Another major application is neural machine translation with translation memory, where the memory can either be the bilingual training corpus \cite{feng-etal-2017-memory,gu2018search} or a large collection of monolingual corpus \cite{cai-etal-2021-neural}.
It also has received great attention in other text generation tasks including dialogue response generation \cite{cai-etal-2019-skeleton,li-etal-2021-retrieve} and knowledge-intensive generation \cite{lewis2020retrieval}, as well as some information extraction tasks including named entity recognition \cite{wang-etal-2021-improving}, and relation extraction \cite{sigir21-retre-re}.

\section{Conclusion}
We tackle Contract Clause Extraction by exploring three implicit relations in contracts. We comprehensively analyze the complexities of contracts and distill out three implicit relations. Then we propose a framework {\tt ConReader} to effectively exploit these relations for solving CCE in complex contracts. Extensive Experiments show that {\tt ConReader} makes considerable improvements over existing methods on two CCE tasks in both conventional and zero-shot settings. Moreover, our analysis towards interpretability also demonstrates that {\tt ConReader} is capable of identifying the supporting knowledge that aids in clause extraction.

\section*{Limitations}
In this section, we discuss the limitations of this work as follows:
\begin{itemize}
    \item In this paper, we employ some language-dependent methods to extract the definitions. Specifically, we use some regular expressions to extract definitions from English contracts in the TDR solver due to the well-organized structure of contracts.
    Therefore, some simple extraction methods have to be designed to tackle the definition extraction when applying our framework to legal contracts in other languages.
    \item In order to meet the need of the end users, there is much room for improvement of the CCE models.
    Due to the limited training data from \textbf{CUAD} (408 contracts), it would be difficult to train a robust model that can be directly used in real-life applications, especially those requiring the zero-shot transfer capability. 
    Therefore, it would be beneficial to collect more training data in order to satisfy the industrial requirements. In addition, the low-resource setting is also a promising and practical direction for future studies.
\end{itemize}
\section*{Ethics Statement}
The main purpose of CCE is to reduce the tedious search effort of legal professionals from finding needles in a haystack. It only serves to highlight potential clauses for human attention and the legal professionals still need to check the quality of those clauses before continuing to the final contract review (still human work). In fact, we use P@0.8R as one of our evaluation metrics because it is quite strict and meets the need of legal professionals. We also conduct a zero-shot setting experiment to demonstrate that the benefit of {\tt ConReader} is not learning from biased information and has a good generalization ability.

We use publicly available CCE corpora to train and evaluate our {\tt ConReader}. The parties in these contracts are mostly companies, which do not involve gender or race issues. 
Some confidential information has originally been redacted to protect the confidentiality of the parties involved. Such redaction may show up as asterisks (***) or underscores (\_\_\_) or blank spaces. 
We make identify and annotate all definitions in those contracts. Such definitions are well structured, which require little legal knowledge. These annotations are just to verify the effectiveness of TDR Solver in {\tt ConReader} but not to contribute a new dataset. We can release the annotated definitions for the reproduction of our analysis if necessary.
We report all pre-processing procedures, hyper-parameters, evaluation schemes, and other technical details and will release our codes for reproduction (we move some to the Appendix due to the space limitation).

\bibliography{anthology,custom}
\bibliographystyle{acl_natbib}

\appendix

\clearpage
\section{Appendix}

\subsection{Data Statistics}
\label{sec:data}
We show the datasets statistics in Table \ref{tab:data}. 
CUAD annotates 41 types of clauses that lawyers need to pay attention to when reviewing contracts. Some types are "Governing Law", "Agreement Date", "License Grant", and "Insurance" et al.
Contract Discovery annotates another 21 types of clauses that must be well-understood by the legal annotators. These types include "Trustee Appointment", "Income Summary", and "Auditor Opinion" et al.
The two datasets differ substantially in their annotated types, making Contract Discovery a good resource for conducting zero-shot experiments.
To prepare a real zero-shot setting, we further remove 6 types of clauses annotated in both corpora to prepare a real zero-shot setting. The types include: \textit{change of control covenant, change of control notice, governing law, no solicitation, effective date reference, effective date main}. 

Since most contents in contracts are unlabeled, which cause a large imbalance between extractable and non-extractable segments.
If a CCE model is trained on this imbalanced data, it is likely to output an empty span since it has been taught by the non-extractable segments not to extract clauses. 
Therefore, we follow \citet{hendrycks2021cuad} to downweight contract segments that do not contain any relevant clauses in the training set such that extractable and non-extractable segments are approximately balanced (i.e. 1:1). While in test sets, we keep all non-extractable segments. This explains why test sets have fewer contracts but more segments.

\begin{table*}[t]
    \centering
    \begin{tabular}{@{}l|c|c|c||ccc@{}} \toprule
    Task & Source &\#Type &Dataset    & \#Contract  & \#Segment &  \#Clause \\\midrule
    \multirow{4}{*}{CA} & CUAD & 41 &Train & 408 & 38,226   & 11,180\\
    & CUAD & 41 &Test (Conv.)    & 102 & 155,098  & 2,643\\
    & Contract Discovery & 15 &Dev (Zero.)     & 287 & 407,907  & 1,031\\
    & Contract Discovery & 15 &Test (Zero.)    & 286 & 375,606  & 1,031\\\midrule
    \multirow{4}{*}{CD} & CUAD & 41 &Train & 408 & 55,249   & 15,988\\
    & CUAD &41 &Test (Conv.)    & 102 & 711,282  & 10,448\\
    & Contract Discovery & 15 &Dev (Zero.)     & 287 & 602,236  & 5,524 \\
    & Contract Discovery & 15 &Test (Zero.)    & 286 & 560,721  & 5,549 \\ \bottomrule
    \end{tabular}
    \caption{Dataset statistics for CA and CD.}
    \label{tab:data}
\end{table*}

\subsection{Annotation Difference}
\label{sec:anno}
Table~\ref{tab:annotation} shows the annotation difference between CUAD and Contract Discovery on ``Governing Law" clauses. In fact, Contract Discovery tends to annotate more facts into the clause, such as parties' obligations. Due to such annotation difference, we also regard an extracted clause as true positive in calculating AUPR if it is a sub-string of the ground truth in the zero-shot setting.

\subsection{Performance by Type}
\label{sec:type}
\begin{figure*}[t]
    \centering
    \includegraphics[scale=1.1]{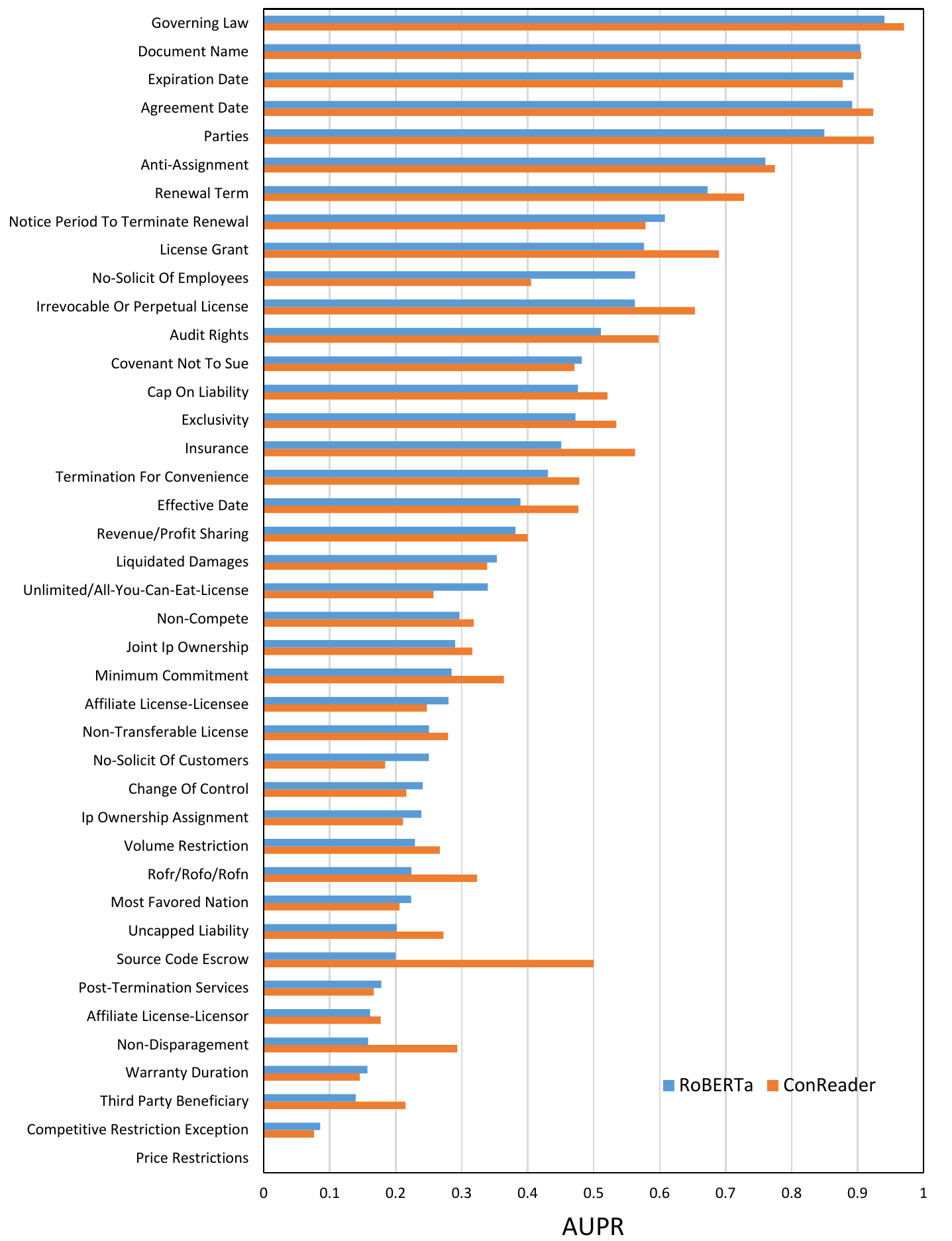}
    \caption{CA performance (AUPR) by clause types. }
    \label{fig:froup}
\end{figure*}

Figure~\ref{fig:froup} shows the AUPR scores for each clause type of {\tt ConReader} and RoBERTa.

\begin{table*}[]
    \centering
    \begin{tabular}{|c|p{148mm}|} \hline
       \multirow{8}{*}{\rotatebox{90}{ \em  CUAD}} & This Agreement shall be construed in accordance with and governed by the substantive internal laws of the State of New York.\\ \cline{2-2}
        & This Agreement shall be governed by the laws of the State of New York, without giving effect to its principles of conflicts of laws, other than Section 5-1401 of the New York General Obligations Law.\\ \cline{2-2}
        & This Agreement is subject to and shall be construed in accordance with the laws of the Commonwealth of Virginia with jurisdiction and venue in federal and Virginia courts in Alexandria and Arlington, Virginia.\\\hline \hline
       \multirow{17}{*}{\rotatebox{90}{ \em  Contract Discovery}} & Section 4.8 Choice of Law/Venue . This Agreement will be governed by and construed and enforced in accordance with the internal laws of the State of California, without giving effect to the conflict of laws principles thereof. Each Party hereby submits to personal jurisdiction before any court of proper subject matter jurisdiction located in Los Angeles, California, to enforce the terms of this Agreement and waives any and all objections to the jurisdiction and proper venue of such courts.\\\cline{2-2}
        & This Agreement will be governed by and 4 construed in accordance with the laws of the State of Delaware (without giving effect to principles of conflicts of laws). Each Party: (a) irrevocably and unconditionally consents and submits to the jurisdiction of the state and federal courts located in the State of Delaware for purposes of any action, suit or proceeding arising out of or relating to this Agreement;\\\cline{2-2}
        & Section 4.8. Choice of Law/Venue . This Agreement will be governed by and construed and enforced in accordance with the internal laws of the State of California, without giving effect to the conflict of laws principles thereof. Each Party hereby submits to personal jurisdiction before any court of proper subject matter jurisdiction located in Los Angeles, California, to enforce the terms of this Agreement and waives any and all objections to the jurisdiction and proper venue of such courts.\\\hline
    \end{tabular}
    \caption{Examples of annotation of ``Governing Law“ clauses in two datasets.}
    \label{tab:annotation}
\end{table*}

\end{document}